\documentclass[journal]{IEEEtran}

\usepackage{graphicx}
\usepackage{subfigure}
\usepackage{url}
\usepackage{times}
\usepackage{amssymb}
\usepackage{array}
\usepackage[ruled,lined]{algorithm2e}
\usepackage{amsmath}
\usepackage{booktabs}
\usepackage{multirow}
\usepackage{cite}
\usepackage{enumerate}

\begin{document}

\title{Parameter Sharing Exploration and Hetero-center Triplet Loss for Visible-Thermal Person Re-Identification}

\author{Haijun Liu, Xiaoheng Tan and Xichuan Zhou

\thanks{H. Liu, X. Tan and X. Zhou are with the School of Microelectronics and Communication Engineering, Chongqing University, Chongqing, 400044, China. (Corresponding author: haijun\_liu@126.com)}}

\markboth{}
{Liu \MakeLowercase{\textit{et al.}}: }

\maketitle

\begin{abstract}
This paper focuses on the visible-thermal cross-modality person re-identification (VT Re-ID) task, whose goal is to match person images between the daytime visible modality and the nighttime thermal modality. The two-stream network is usually adopted to address the cross-modality discrepancy, the most challenging problem for VT Re-ID, by learning the multi-modality person features. In this paper, we explore how many parameters a two-stream network should share, which is still not well investigated in the existing literature. By splitting the ResNet50 model to construct the modality-specific feature extraction network and modality-sharing feature embedding network, we experimentally demonstrate the effect of parameter sharing of two-stream network for VT Re-ID.
Moreover, in the framework of part-level person feature learning, we propose the hetero-center triplet loss to relax the strict constraint of traditional triplet loss by replacing the comparison of the \emph{anchor to all the other samples} by the \emph{anchor center to all the other centers}.
With extremely simple means, the proposed method can significantly improve the VT Re-ID performance.
The experimental results on two datasets show that our proposed method distinctly outperforms the state-of-the-art methods by large margins, especially on the RegDB dataset achieving superior performance, rank1/mAP/mINP 91.05\%/83.28\%/68.84\%. It can be a new baseline for VT Re-ID, with a simple but effective strategy.

\end{abstract}


\section{Introduction}
\label{sec:intro}
\IEEEPARstart{P}{erson} re-identification (Re-ID) can be regarded as a retrieval task, which aims at searching a person of interest from multi-disjoint cameras deployed at different locations. It has received increasing interest in the computer vision community due to its importance in intelligent video surveillance and criminal investigation applications. Visible-visible Re-ID (VV Re-ID), the most common single-modality Re-ID task, has progressed and achieved high performance in recent years \cite{Ye2020DeepLF}.

\begin{figure}
\centering
\includegraphics[width=9cm]{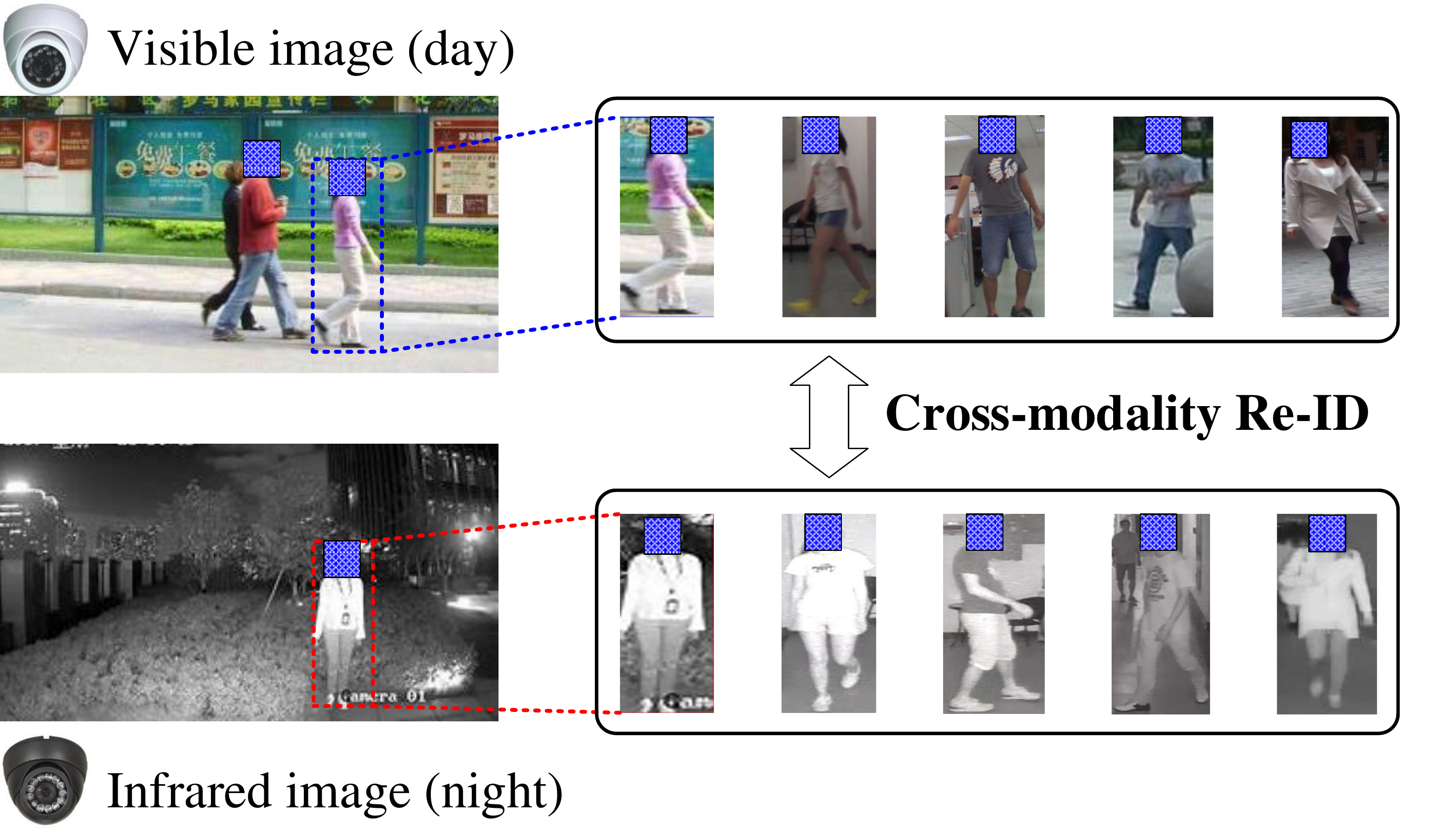}
\caption{Illustration of VT Re-ID. For example, searching a person captured by a visible camera in the daytime among multiple persons captured by infrared (or thermal) cameras at night, and vice versa. }
\label{fig:vt_reid_illus}
\end{figure}

However, in practical scenarios, a 24-hour intelligent surveillance system, the visible-thermal cross-modality person re-identification (VT Re-ID) problem, is frequently encountered, as shown in Fig. \ref{fig:vt_reid_illus}. For example, criminals always collect information during the day and execute crimes at night, in which case, the query image may be obtained from the thermal camera (or the infrared camera) during the nighttime, while the gallery images may be captured by the visible cameras during the daytime.

In recent years, an increasing number of researchers have focused on the VT Re-ID task, achieving substantial progresses with novel and effective ideas. However, many works evaluated the effectiveness of their methods with a poor baseline, which seriously impedes the development of the VT Re-ID community. \textbf{In the present study, our proposed method can be set as a strong and effective baseline for VT Re-ID with some extremely simple means.}

The VT Re-ID task suffers from two major problems, the large \textbf{cross-modality discrepancy} arisen from the different reflective visible spectra and sensed emissivities of visible and thermal cameras, and the large \textbf{intra-modality variations}, similar to the VV Re-ID task, caused by viewpoint changing and different human poses, etc.
To alleviate the extra cross-modality discrepancy in VT Re-ID, an intuitive and apparent method is to map the cross-modality persons into a common feature space to realize the similarity measure. Therefore, a two-stream framework is always adopted, including two modality-specific networks with independent parameters for feature extraction, and a parameter-sharing network for feature embedding to project the modality-specific features into a common feature space.
Generally, the two modality-specific networks are not required to have the same architecture. The only criterion is that their outputs should have the same dimension shapes to be the input of the parameter-sharing network for feature embedding.
In the literature, the ResNet50 \cite{he2016deep} model is preferentially adopted as the backbone to construct the two-stream network, all the res-convolution blocks for feature extraction and some parameter-sharing fully connected layers for feature embedding. However, is this setting the best choice to construct the two-stream network? Those parameter-sharing fully-connected layers can only process the 1D-shaped vector, ignoring the spatial structure information of persons. To take advantage of the convolutional layers for processing the 3D-shaped tensor with spatial structure information, we can share some parameters of res-convolution blocks for feature embedding. In this situation, \textbf{how many parameters of the two-stream network to share is the point of this study to investigate.}

In addition, the network is always trained with identification loss and triplet loss to simultaneously enlarge the inter-class distance and minimize the intra-class distance.
The triplet loss is performed on \emph{each anchor sample to all the other samples} from both the same modality and cross modality. This may be a strong constraint for constraining the pairwise distance of those samples, especially when there exist some outliers (bad examples), which would form the adverse triplet to destroy other well-learned pairwise distances. It also leads to high complexity with a large number of triplets.
The cross-modality and intra-modality training strategy is separately employed to enhance feature learning \cite{ye2018visible,liu2020enhancing}. In the authors' opinion, the separate cross-modality and intra-modality training strategy may be unnecessary if those learned person features by the two-stream network are good enough in the common feature space, where the features can hardly be distinguished from which modality.
\textbf{Therefore, we propose the hetero-center triplet loss that directly performs in the unified common feature space.} The hetero-center triplet loss is performed on \emph{each anchor center to all the other centers}, which can also reduce the computational complexity.

The main contributions can be summarized as follows.
\begin{itemize}
\item We achieve state-of-the-art performance on two datasets by large margins, which can be a strong VT Re-ID baseline to boost future research with high quality.
\item We explore the parameter-sharing problem in a two-stream network. To the best of our knowledge, it is the first attempt to analyze the impact of the number of parameter sharing for cross-modality feature learning.
\item We propose the hetero-center triplet loss to constrain the distance of different class centers from both the same modality and cross modality.
\end{itemize}

\section{Related work}
\label{sec:relatedwork}
This section briefly reviews those existing VT Re-ID approaches. Compared to the traditional VV Re-ID, except for the intra-modality variations, VT Re-ID should handle the extra cross-modality discrepancy. To alleviate this problem, researchers focus on projecting (or translating) the heterogeneous cross-modality person images into a common space for similarity measure, mainly including the following aspects: network designing, metric learning and image translation.

\subsection{Network designing}
Feature learning is the fundamental step of Re-Identification before similarity measure. Most studies focus on the visible and thermal person feature learning through designing deep neural networks (DNN).
Ye et.al \cite{ye2018hierarchical,ye2018visible,tifs19vtreid} proposed adopting a two-stream network to separately extract the modality-specific features, and then performed the feature embedding to project those features into the common feature space with parameters sharing fully connected layers.
Based on the two-stream network, Liu et al. \cite{liu2020enhancing} introduced mid-level features incorporation to enhance the modality-shared person features with more discriminability.
To learn good modality-shared person features, Dai et al. \cite{dai2018cross} proposed the cross-modality generative adversarial network (cmGAN) under the adversarial learning framework, including a discriminator to distinguish whether the input features are from the visible modality or thermal modality.
Zhang et al. \cite{zhang2019attend} proposed a dual-path cross-modality feature learning framework, including a dual-path spatial-structure-preserving common space network and a contrastive correlation network, which preserves intrinsic spatial structures and attends to the difference of input cross-modality image pairs.
To explore the potential of both the modality-shared information and the modality-specific characteristics to boost the re-identification performance, Lu et al. \cite{lu2020cross} proposed modeling the affinities of different modality samples according to the shared features and then transferring both shared and specific features among and across modalities.

Moreover, for handling the cross-modality discrepancy, some works concentrate on the input design of a single-stream network to simultaneously utilize visible and thermal information.
Wu et al. \cite{wu2017rgb} first proposed to study the VT Re-ID problem, built the SYSU-MM01 dataset, and developed the zero-padding method to extract the modality-shared person features with a single-stream network.
Kang et al. \cite{kang2019person} proposed combining the visible and thermal images as a single input with different image channels.
Additionally, Wang et al. \cite{wang2019learning1} also adopted a multispectral image as the input for feature learning, where the multispectral image consists of the visible image and corresponding generated thermal image, or the generated visible image and corresponding thermal image.

\subsection{Metric learning}
Metric learning is the key step of Re-ID for similarity measure. In the deep learning framework, due to the advantage of DNN on feature learning, Re-ID could achieve good performance with only the Euclidean distance metric. Therefore, metric learning is inherent in the training loss function of DNN, guiding the training process to make the extracted features more discriminative and robust.
Ye et al. \cite{ye2018hierarchical} proposed a hierarchical cross-modality matching model by jointly optimizing the modality-specific and modality-shared metrics in a sequential manner. Then, they presented a bi-directional dual-constrained top-ranking loss to learn discriminative feature representations based on a two-stream network \cite{ye2018visible}, based on which, the center-constraint was also introduced to improve performance \cite{tifs19vtreid}.
Zhu et al. \cite{zhu2019hetero} proposed the hetero-center loss to reduce the intra-class cross-modality variations.
Liu et al. \cite{liu2020enhancing} also proposed the dual-modality triplet loss to guide the training procedures by simultaneously considering the cross-modality discrepancy and intra-modality variations.
Hao et al. \cite{hao2019hsme} proposed an end-to-end two-stream hypersphere manifold embedding network with both classification and identification loss, constraining the intra-modality variations and cross-modality variations on this hypersphere.
Zhao et al. \cite{zhao2019hpiln} introduced the hard pentaplet loss to improve the performance of the cross-modality re-identification.
Wu et al. \cite{wu2020rgb} cast the learning shared knowledge for cross-modality matching as the problem of cross-modality similarity preservation, and proposed a focal modality-aware similarity-preserving loss to leverage the intra-modality similarity to guide the inter-modality similarity learning.

\subsection{Image translation}
The aforementioned works handle the cross-modality discrepancy and intra-modality variations from the feature extraction level. Recently, image generation methods based on generative adversarial network (GAN) have drawn much attention in VT Re-ID, reducing the domain gap between visible and thermal modalities from image level.
Kniaz et al. \cite{kniaz2018thermalgan} first introduced GAN to translate a single visible image to a multimodal thermal image set, and then performed the Re-ID in the thermal domain.
Wang et al. \cite{wang2019rgb} proposed an end-to-end alignment generative adversarial network (AlignGAN) for VT Re-ID, to jointly bridge the cross-modality gap with feature alignment and pixel alignment.
Wang et al. \cite{wang2019learning1} proposed a dual-level discrepancy reduction learning framework based on a bi-directional cycleGAN to reduce the domain gap, from both the image and feature levels.
Choi et al. \cite{choi2020hi} proposed a hierarchical cross-modality disentanglement (Hi-CMD) method, which automatically disentangles ID-discriminative factors and ID-excluded factors from visible-thermal images. Hi-CMD includes an ID-preserving person image generation network and a hierarchical feature learning module.

However, a person in the thermal modality can have different colors of clothes in the visible modality, leading to one thermal person image corresponding to multiple reasonable visible person images by image generation. It is hard to know which one is the correct target to be generated for Re-ID since when generating images, the model cannot access the gallery images that only appear in the inference phase.
Image generation-based methods always have performance uncertainty, high complexity and high training trick demands.

\section{Our proposed method}
\label{sec:method}
\begin{figure*}
\centering
\includegraphics[width=18cm]{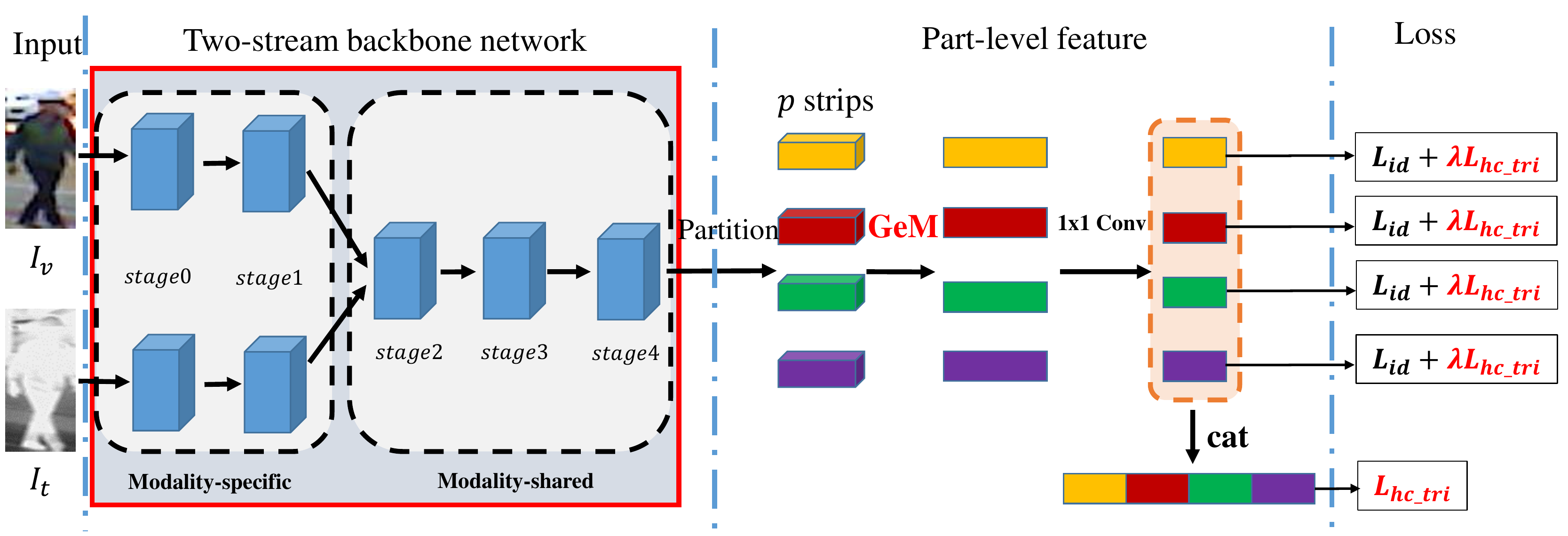}
\caption{The pipeline of our proposed framework for VT Re-ID, which mainly contains three components: two-stream backbone network, part-level feature learning block and loss. The two-stream backbone network includes two modality-specific branches with independent parameters and follows one modality-shared branch with shared parameters. For example, we take the ResNet50 model as the backbone, the first two stages ($stage0$ and $stage1$) form the modality-specific branches and the following three stages ($stage2$, $stage3$ and $stage4$) form the modality-shared branch.
Then, the feature map outputted from the backbone is horizontally split into $p$ 3D tensors, here $p=4$, which are pooled into vectors by generalized-mean ($GeM$) pooling operation. For each part vector, a $1 \times 1$ Conv block reduces the dimension of features. Afterward, the reduced part features are respectively input to compute the identification loss $L_{id}$ and our proposed hetero-center triplet loss $L_{hc\_tri}$. Finally, all the part features are concatenated ($cat$) to form the final person features, which is supervised by $L_{hc\_tri}$. }
\label{fig:framework}
\end{figure*}
In this section, we introduce the framework of our proposed feature learning model for VT Re-ID, as depicted in Fig. \ref{fig:framework}. The model mainly consists of three components: (1) the two-stream backbone network, exploring the parameter sharing, (2) the part-level feature extraction block and (3) the loss, our proposed hetero-center triplet loss and identity softmax loss.

\subsection{Two-stream backbone network}
\label{ssec:backbone}
The two-stream network is a conventional way to extract features in visible-thermal cross-modality person re-identification, first introduced in \cite{ye2018visible}. It consists of two parts: feature extractor and feature embedding. The feature extractor aims at learning the modality-specific information from two heterogeneous modalities, while the feature embedding focuses on learning the multi-modality shared features for cross-modality re-identification by projecting those modality-specific features into a modality-shared common feature space.

In the existing literature, feature embedding is always computed by some shared fully connected layers, and the feature extractor is always a well-designed convolution neural network, such as ResNet50. In this situation, there may be two problems we should pay attention.
\begin{enumerate}
\item The feature extractor consists of two branches with independent parameters. If each branch consists of the whole well-designed CNN architecture, the number of network parameters (model size) would increase by double.
\item Feature embedding consists of some shared fully connected layers, which can only process the 1D-shaped feature vector without any person spatial structure information. However, the person spatial structure information is crucial to describe a person.
\end{enumerate}

\begin{table}
\caption{Different splits of ResNet50 model to form the two-stream backbone network. $\phi_{v}$ and $\phi_{t}$ respectively denote the visible-stream and thermal-stream feature extraction network. $\phi_{vt}$ denotes the modality-shared feature embedding network. $stage\{0-2\}$ denotes $stage0$, $stage1$ and $stage2$. It is detailedly illustrated in Fig. \ref{fig:net_share}.}
\label{tab:split_scheme}
  \centering
  \begin{tabular}{l|c|c}
  \toprule[2pt]
       \multirow{3}{*}{} & Modality-specific  & Modality-shared  \\
        & feature extractor & feature embedding \\
       & ($\phi_{v}$ and $\phi_{t}$) & ($\phi_{vt}$) \\ \toprule[1pt]
      $s0$ & - & $stage\{0-4\}$  \\ \hline
      $s1$ & $stage\{0\}$ & $stage\{1-4\}$  \\ \hline
      $s2$ & $stage\{0-1\}$ & $stage\{2-4\}$  \\ \hline
      $s3$ & $stage\{0-2\}$ & $stage\{3-4\}$  \\ \hline
      $s4$ & $stage\{0-3\}$ & $stage\{4\}$  \\ \hline
      $s5$ & $stage\{0-4\}$ & -  \\
     \toprule[2pt]
  \end{tabular}
\end{table}

To simultaneously address the aforementioned two problems, we propose to split the well-designed CNN model into two parts. The former part can be set as a two-stream feature extractor with independent parameters, while the latter part can be set as the feature embedding model. In this way, the whole model size will be reduced (corresponding to problem 1). The input of the feature embedding block is the output of the feature extractor, only the middle 3D feature maps of the well-designed CNN model, which is full of the person spatial structure information (corresponding to problem 2).

Therefore, the key point is how to split the well-designed CNN model. Namely, how many parameters of the two-stream network should be independent to learn the modality-specific information?

For simplicity in presentation, we denote the visible-stream feature extraction network as function $\phi_{v}$, the thermal-stream feature extraction network as $\phi_{t}$ to learn the modality-specific information, and the feature embedding network as $\phi_{vt}$ to project modality-specific person features into the shared common feature space.
Given a visible image $I_{v}$ and a thermal image $I_{t}$, the learned 3D person features $v$ and $t$ in common space can be represented as,
\begin{align}
\left\{
    \begin{tabular}{cl}
    $v = \phi_{vt}\big( \phi_{v}(I_{v}) \big)$,\\
    $t = \phi_{vt}\big( \phi_{t}(I_{t}) \big)$.
    \end{tabular}
    \right.
\end{align}

We adopt the ResNet50 model as the backbone, with the consideration of its competitive performance in some Re-ID systems as well as its relatively concise architecture. The ResNet50 model mainly consists of one shallow convolution block $stage0$ and four res-convolution blocks, $stage1$, $stage2$, $stage3$ and $stage4$.
To split the ResNet50 model into our modality-specific feature extractor and modality-shared feature embedding network, we can sequentially obtain the split scheme as shown in Table \ref{tab:split_scheme}, where $si, i=\{0,1,2,3,4,5\}$ means $\phi_{vt}$ starts from the $i^{th}$ $stage$. $s0$ and $s5$ are two extreme cases. $s0$ means that the two-stream backbone network shares all the ResNet50 model without the modality-specific feature extractor, while $s5$ means that all parameters of the two streams for the visible and thermal modalities are totally independent as in \cite{ye2018visible}.
Which is the best choice for a two-stream backbone network for cross-modality Re-ID? In the authors' opinion, these two extreme cases $s0$ and $s5$ are not good, since they ignore some important information in the cross-modality Re-ID task. Experimental results in Sec. \ref{sssec:banckbone} show that the modality-shared feature embedding network comprising some res-convolution blocks is a good choice, since the input of modality-shared feature embedding network $\phi_{vt}$ is 3D shape feature maps, with the spatial structure information of persons.

\subsection{Part-level feature extraction block}
In VV Re-ID, state-of-the-art results are always achieved with part-level deep features \cite{Yang2020PPU,Wan2020CLPD}. A typical and simple approach is partitioning persons into horizontal strips to coarsely extract the part-level features, which can then be concatenated to describe the person's body structure. Body structure is the inherent characteristic of a person, which is invariant information of the person's body whatever modality the image is captured from. Namely, the body structure information is modality-invariant, which can be adopted as modality-shared information to represent a person.
Therefore, according to the part-level feature extraction method in \cite{sun2018beyond,wang2018learning}, we also adopt the uniform partition strategy to obtain coarse body part features.

Given a person (visible or thermal) images, it will become the 3D feature map after undergoing all the layers inherited from the two-stream backbone network. Based on the 3D feature maps, as shown in Fig. \ref{fig:framework}, there are 3 steps to extract the part-level person features as follows.
\begin{enumerate}
  \item The 3D feature maps are \textbf{uniformly partitioned} into $p$ strips in the horizontal orientation to generate the coarse body part feature maps, as shown in Fig. \ref{fig:framework}, where $p=4$.
  \item Instead of utilizing the widely used max-pooling or average pooling, we adopt a generalized-mean (\textbf{GeM}) \cite{radenovic2018fine} pooling layer to translate the 3D part feature maps into the 1D part feature vectors. Given a 3D feature patch $X \in R^{C \times H \times W}$, the GeM can be formulated as,
      \begin{align}
        \hat{x} = \big(\frac{1}{|X|}\sum_{x_{i}\in X} x_{i}^{\textrm{p}} \big)^{\frac{1}{\textrm{p}}},
      \end{align}
      where $\hat{x} \in R^{C \times 1 \times 1}$ is the pooled result, $|\cdot|$ denotes the element number, and $\textrm{p}$ is the pooling hyperparameter, which can be preset or learned by back-propagation.
  \item Afterwards, a $1 \times 1$ convolutional (\textbf{$1 \times 1$ Conv}) block is employed to reduce the dimension of part-level feature vectors. The block includes a $1 \times 1$ convolutional layer whose output channel number is $d$, following a batch normalization layer and a ReLU layer.
\end{enumerate}

Moreover, each part-level feature vector is first adopted to perform metric learning with triplet loss $L_{tri}$ (or our proposed hetero-center triplet loss $L_{hc\_tri}$). Then, a fully connected layer with desired dimensions (corresponding to the number of identities in our model) is adopted to perform the identification with softmax $L_{id}$. There are $p$ part-level features that need $p$ different classifiers without sharing parameters.

Finally, all the $p$ part-level features are concatenated ($cat$) to form the final person features for the similarity measure during testing. Additionally, the final person features can also be supervised by the $L_{tri}$ (or $L_{hc\_tri}$).

\begin{figure}
\centering
\includegraphics[width=9cm]{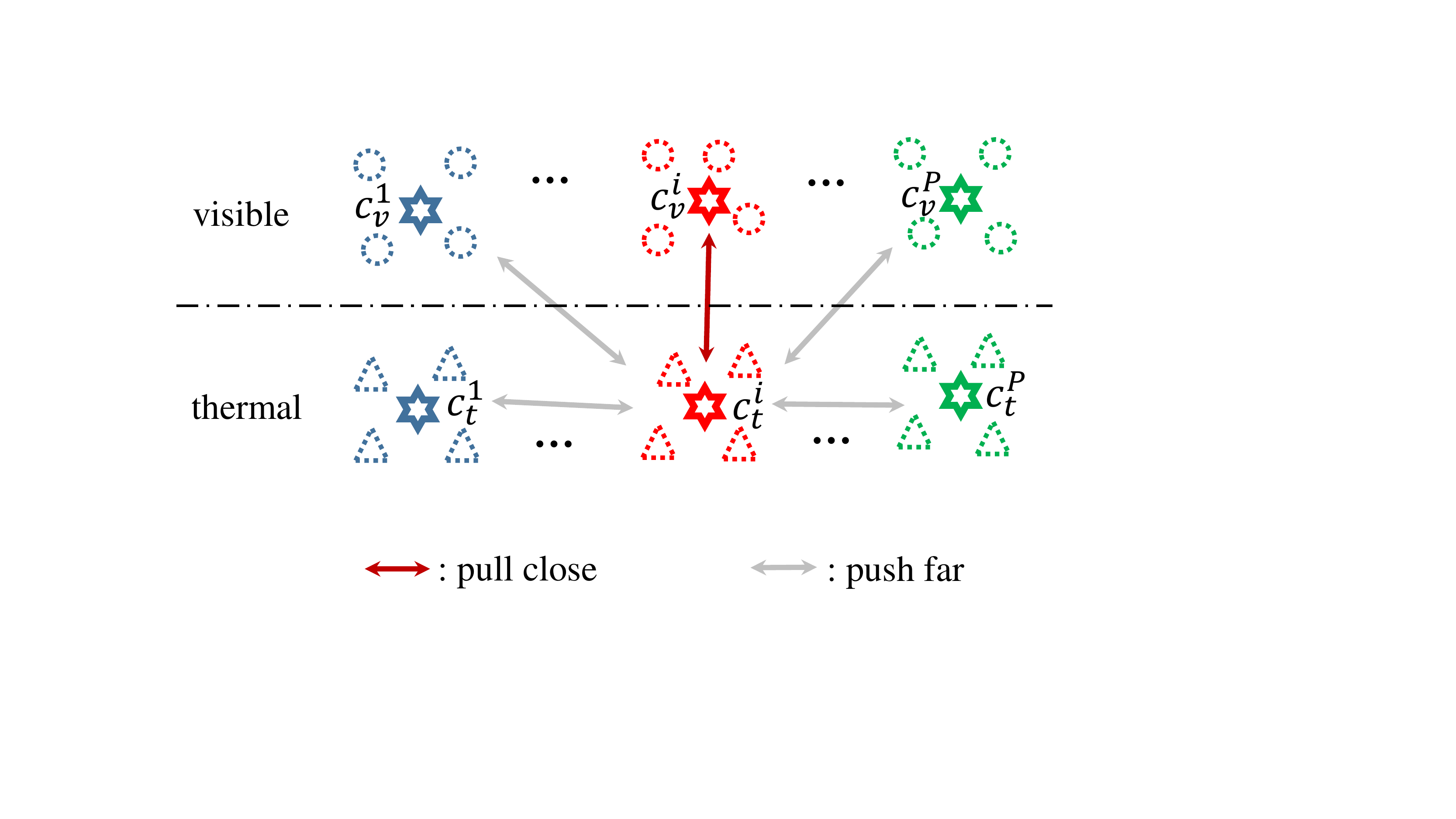}
\caption{Illustration of the hetero-center triplet loss, which aims pulling close those centers with the same identity label from different modalities, while pushing far away those centers with different identity labels regardless of which modality it is from. We compare the \emph{center to center similarity} rather than \emph{sample to sample similarity} or \emph{sample to center similarity}. The stars denote the centers. Different colors denote different identities. }
\label{fig:center_illus}
\end{figure}

\subsection{The hetero-center triplet loss}
In this subsection, we introduce the designed hetero-center triplet loss to guide the network training for part-level feature learning. The learning objective is directly conducted in the common feature space to simultaneously deal with both cross-modality discrepancy and intra-modality variations.
First, we revisit the general triplet loss.

\subsubsection{Triplet loss revisit}
Triplet loss was first proposed in FaceNet \cite{schroff2015facenet}, and then improved by mining the hard triplets \cite{hermans2017defense}. The core idea is to form batches by randomly sampling $P$ identities, and then randomly sampling $K$ images of each identity, resulting in a mini-batch with $PK$ images. For each sample $x_a$ in the mini-batch, we can select the hardest positive and hardest negative samples within the mini-batch to form the triplets for computing the batch hard triplet loss,
\begin{align}\label{eq:loss_bh}
    L_{bh\_tri}(X) = \overbrace{\sum\limits_{i=1}^{P} \sum\limits_{a=1}^{K}}^{\textnormal{all anchors}}
        \Big[\rho & + \hspace*{-5pt} \overbrace{\max\limits_{p=1 \dots K} \hspace*{-5pt} \|x^i_a - x^i_p\|_2 }^{\textnormal{hardest positive}} \\
               & - \hspace*{-5pt} \underbrace{\min\limits_{\substack{j=1 \dots P \\ n=1 \dots K \\ j \neq i}} \hspace*{-5pt} \| x^i_a - x^j_n \| _2}_{\textnormal{hardest negative}} \Big]_+,\nonumber
\end{align}
which is defined for a mini-batch $X$, where a data point $x_a^i$ denotes the $a^{th}$ image of the $i^{th}$ person in the batch, $[x]_{+} = \max(x, 0)$ denotes the standard hinge loss, $\| x_a - x_p \|_2$ denotes the Euclidean distance of data point $x_a$ and $x_p$, $\rho$ is the margin.

\subsubsection{Batch sampling method} Due to our two-stream structure respectively extracting features for visible and thermal images, we introduce the following online batch sampling strategy. Specifically, $P$ person identities are first randomly selected at each iteration, and then we randomly select $K$ visible images and $K$ thermal images of the selected identity to form the mini-batch, in which a total of $2*PK$ images are obtained. This sampling strategy can fully utilize the relationship of all the samples within a mini-batch. In this manner, the sample size of each class is the same, which is important to avoid the perturbations caused by class imbalance. Moreover, due to the random sampling mechanism, the local constraint in the mini-batch can achieve the same effect as the global constraint in the entire set.

\begin{table}
  \centering
  \caption{The comparison of computational cost between general triplet loss $L_{bh\_tri}$ and our proposed hetero-center triplet loss $L_{hc\_tri}$. Due to the symmetrical property of distance measure, the computational cost could divide 2. }\label{tab:cost_com}
  \begin{tabular}{l|c|c}
  \toprule[2pt]
     & positive   & negative  \\ \toprule[1pt]
   $L_{bh\_tri}$ &      $ 2PK \times (2K-1)$    &  $2PK \times 2(P-1)K $        \\ \hline
   $L_{hc\_tri}$ &       $2P$          &    $2P \times 2(P-1)$       \\ \toprule[2pt]
  \end{tabular}
\end{table}

\subsubsection{Hetero-center triplet loss} Eq. (\ref{eq:loss_bh}) shows that triplet loss computes the loss by comparison of the \emph{anchor to all the other samples}. It is a strong constraint, perhaps too strict to constrain the pairwise distance if there exist some outliers (bad examples), which would form the adverse triplet to destroy other pairwise distances. Therefore, we consider adopting the center of each person as the identity agent. In this manner, we can relax the strict constraint by replacing the comparison of the \emph{anchor to all the other samples} by the \emph{anchor center to all the other centers}.

First, in a mini-batch, the center for the features of every identity from each modality is computed,
\begin{align}\label{eq:center}
  c_{v}^{i} = \frac{1}{K} \sum_{j=1}^{K} v_{j}^{i},  \\
  c_{t}^{i} = \frac{1}{K} \sum_{j=1}^{K} t_{j}^{i},  \nonumber
\end{align}
which is defined for a mini-batch, where $v_j^i$ denotes the $j^{th}$ visible image feature of the $i^{th}$ person in the mini-batch, while $t_j^i$ corresponds to the thermal image feature.

Therefore, based on our $PK$ sampling method, in each mini-batch, there are $P$ visible image centers \{$c_{v}^{i}| i=1,\cdots,P$\} and $P$ thermal centers \{$c_{t}^{i}| i=1,\cdots,P$\}, as shown in Fig. \ref{fig:center_illus}. In the following, all the computations are only performed on the centers.

The goal of metric learning is to make those features from the same class close to each other (intra-class compactness), while those features from different classes are far away from each other (inter-class separation). Therefore, in our VT cross-domain Re-ID, based on the $PK$ sampling strategy and calculated centers, we can define the hetero-center triplet loss as,
\begin{align}\label{eq:loss_ct}
    L_{hc\_tri}(C) = & \sum\limits_{i=1}^{P}\Big[\rho  +  \|c^i_v - c^i_t\|_2  - \min\limits_{\substack{ n \in \{v, t\} \\ j \neq i}}  \| c^i_v - c^j_n \| _2 \Big]_+  \\
    & + \sum\limits_{i=1}^{P}\Big[\rho  +  \|c^i_t - c^i_v\|_2  - \min\limits_{\substack{ n \in \{v, t\} \\ j \neq i}}  \| c^i_t - c^j_n \| _2 \Big]_+,\nonumber
\end{align}
which is defined on mini-batch centers $C$ including both visible centers \{$c_{v}^{i}| i=1,\cdots,P$\} and thermal centers \{$c_{t}^{i}| i=1,\cdots,P$\}.
For each identity, $L_{hc\_tri}$ concentrates on only one cross-modality positive pair and the mined hardest negative pair in both the intra- and inter-modality.

Comparing the general triplet loss $L_{bh\_tri}$ (Eq. (\ref{eq:loss_bh})) to our proposed center-based triplet loss $L_{hc\_tri}$ (Eq. (\ref{eq:loss_ct})), we replace the comparison of the \emph{anchor to all the other samples} by the \emph{anchor center to all the other centers}. This modification has two major advantages:
\begin{enumerate}[a)]
  \item It reduces the computational cost, as shown in Table \ref{tab:cost_com}. For a mini-batch with $2PK$ images, $L_{bh\_tri}$ requires computing pairwise distance $2PK \times (2K-1)$ for hardest positive sample mining and $2PK \times 2(P-1)K$ for hardest negative sample mining. In comparison, $L_{hc\_tri}$ only needs to compute the pairwise distance $2P$ for positive sample pairs (there are only $P$ cross-modality positive center pairs), and $2P \times 2(P-1)$ for hardest negative center sample mining. The computational cost is largely reduced.
  \item It relaxes the sample-based triplet constraint to the center-based triplet constraint, which also preserves the property of handling both the intra-class and inter-class variations simultaneously on visible and thermal modalities in the common feature space. For each identity, minimizing the only cross-modality positive center pairwise distance can ensure intra-class feature compactness. The hardest negative center mining can ensure the inter-class feature distinguishable property both in visible and thermal modalities.
\end{enumerate}

\begin{figure}
\centering
\footnotesize
\begin{tabular}{c@{\hspace{0mm}}c}
\includegraphics[width=4.3cm]{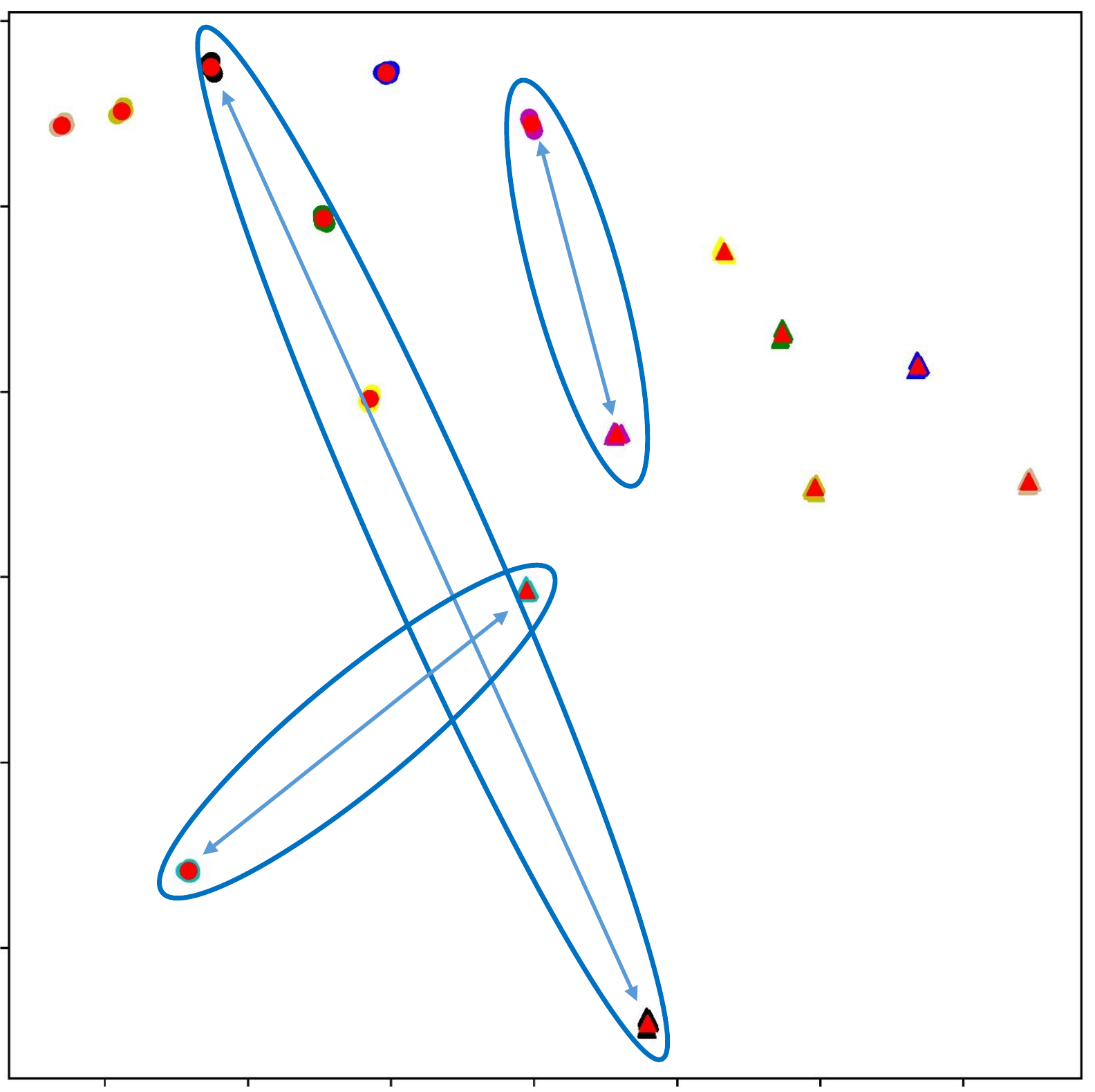} &
\includegraphics[width=4.3cm]{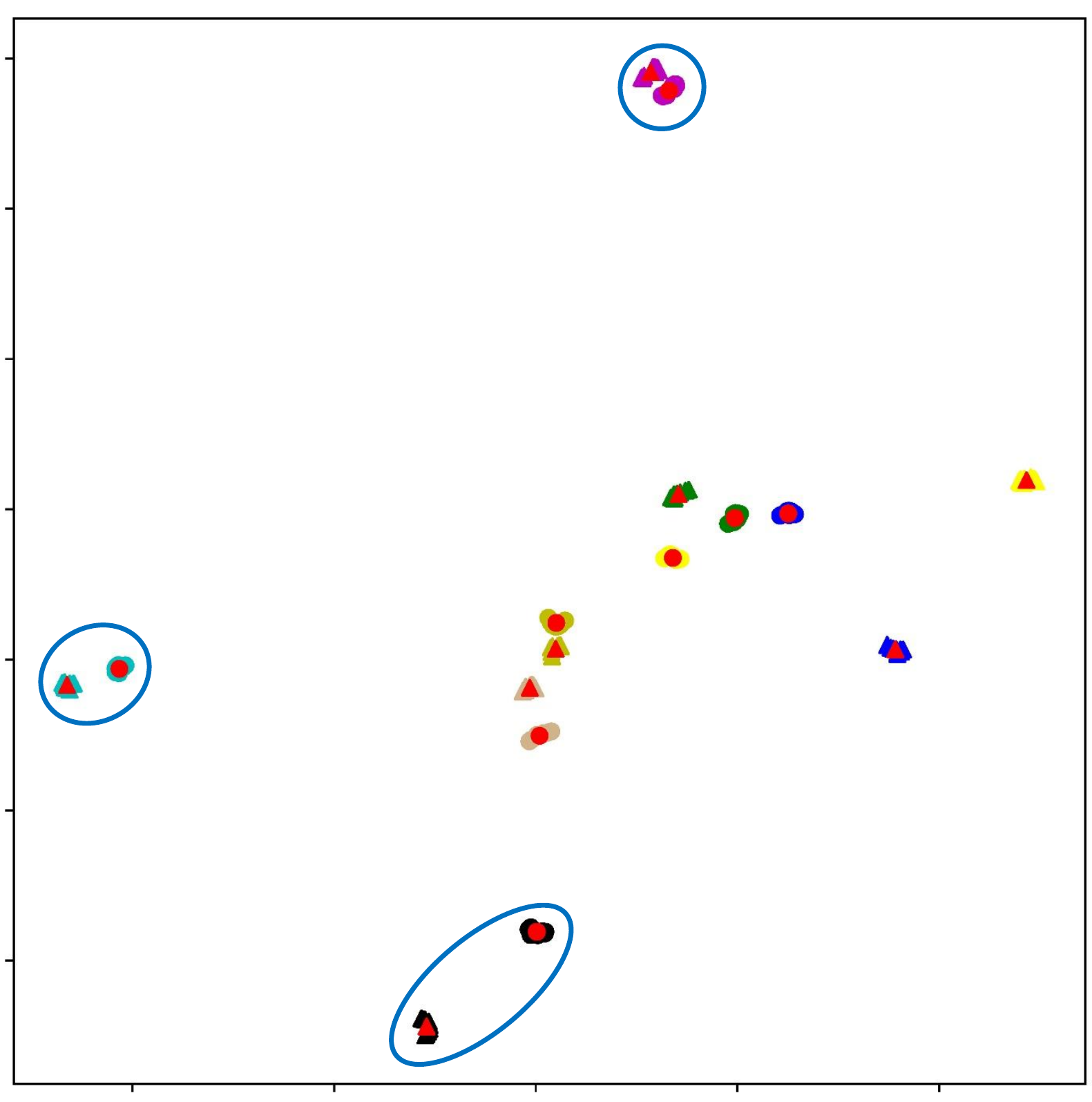}  \\  (a) Learned center loss & (b) Hetero-center triplet loss
\end{tabular}
\caption{The visualization of features extracted by the baseline model with (a) the learned center loss and (b) our proposed hetero-center triplet loss. The features are from 8 randomly chosen identities in the RegDB testing dataset, whose dimension of features is reduced to 2 by tSNE. Points with different colors denote features belonging to different identities. Points of different shapes denote different modalities. The red points with different shapes denote the feature centers of each identity from different modalities. Blue arrows in the blue circles link the two centers of one identity from two modalities.}
\label{fig:visualization}
\end{figure}

\subsubsection{Comparison to other center-based losses}
There are two kinds of center-based losses: the learned centers \cite{Wen2016ADF,tifs19vtreid} and the computed centers \cite{zhu2019hetero}. The main difference lies in the method of obtaining the centers. One learns them by pre-setting a center parameter for each class, while the other computes the centers directly based on the learned deep features.

 \textbf{The learned centers.} The learned center loss \cite{Wen2016ADF} is first introduced in face verification to learn a center for the features of each class and penalizes the distances between the deep features and their corresponding centers. For our cross-modality VT Re-ID task with the $PK$ sampling strategy, the learned center loss can be extended in a bi-directional manner \cite{liu2020enhancing,tifs19vtreid,ye2018visible} as follows,
\begin{align} \label{eq:lc_}
    L_{lc} = \frac{1}{2} \sum_{i=1}^{P} \sum_{j=1}^{K} \big(\| v^{i}_{j} - c^{i} \|_{2} + \| t^{i}_{j} - c^{i} \|_{2}\big),
\end{align}
where $v_j^i$ denotes the $j^{th}$ visible image feature of the $i^{th}$ person in the mini-batch, while $t_j^i$ corresponds to the thermal image feature. $c^{i}$ is the $i^{th}$ class center for both visible and thermal modalities.

Comparing the learned center loss $L_{lc}$ (Eq. (\ref{eq:lc_})) to our proposed hetero-center triplet loss $L_{hc\_tri}$ (Eq. (\ref{eq:loss_ct})), there are the following differences.
1) $L_{hc\_tri}$ is in a comparison of the \emph{anchor center to centers} rather than $L_{lc}$'s \emph{anchor sample to centers}.
2) $L_{hc\_tri}$ computes the centers for visible and thermal modalities, while $L_{lc}$ unifies the $i^{th}$ class center for both visible and thermal modalities into one learned vector.
3) $L_{hc\_tri}$ is formulated by triplet mining the properties of both the inter-class separability and intra-class compactness, while $L_{lc}$ only focuses on the intra-class compactness, ignoring the inter-class separability.

As shown in Fig. \ref{fig:visualization}, our proposed hetero-center triplet loss $L_{hc\_tri}$ concentrates on both of the inter-class separability and intra-class compactness, while the learned center loss $L_{lc}$ ignores the inter-class separability for both the intra- and inter modality.
$L_{lc}$ only performs well on intra-modality intra-class compactness, but poorly on cross-modality intra-class compactness. This may be due to the hard training of learned center loss combined with identification loss, which leads to the unsatisfactory performance.

\textbf{The computed centers.} The other method for obtaining the center of each class is to calculate it directly based on the learned deep features \cite{zhu2019hetero}. We also adopt this approach. Instead of pre-setting a center parameter to be learned as Eq. (\ref{eq:lc_}), the centers are directly calculated as Eq. (\ref{eq:center}).
In \cite{zhu2019hetero}, the hetero-center loss $L_{hc}$ was proposed to improve the intra-class cross-modality similarity, penalizing the center distance between two modality distributions, which can be formulated as follows,
\begin{align} \label{eq:hc_}
    L_{hc} =  \sum_{i=1}^{P} \| c^{i}_{v} - c^{i}_{t} \|_{2}.
\end{align}

Comparing the hetero-center loss $L_{hc}$ (Eq. (\ref{eq:hc_})) to our proposed hetero-center triplet loss $L_{hc\_tri}$ (Eq. (\ref{eq:loss_ct})), the main difference is that $L_{hc}$ only focuses on the intra-class cross-modality compactness (the red arrows in Fig. \ref{fig:center_illus}), while our $L_{hc\_tri}$ additionally focuses on the inter-class separability for both the intra- and inter-modality (the grey arrows in Fig. \ref{fig:center_illus}) with triplet mining. In summary, $L_{hc}$ is only a part of our proposed $L_{hc\_tri}$.

\subsubsection{The overall loss}
Moreover, similar to some state-of-the-art VT Re-ID methods \cite{hao2019hsme,wang2019learning1,ye2018visible,tifs19vtreid,zhu2019hetero}, for the sake of feasibility and effectiveness for classification, identification loss is also utilized to integrate the identity-specific information by treating each person as a class.
The identification loss with label smooth operation is adopted to prevent overfitting the Re-ID model training. Given an image, we denote $y$ as the truth ID label and $p_i$ as the ID prediction logits of the $i^{th}$ class. The identification loss is calculated as follows,
\begin{align}\label{eq:l_id}
  L_{id} &= \sum_{i=1}^{N} -q_{i} \log(p_i) \\
  s.t. & \,\,\,q_i = \left\{
    \begin{tabular}{cl}
    $ 1-\frac{N-1}{N}\xi$, & $y = i$,\\
    $\frac{\xi}{N}$, & $y \neq i$,
    \end{tabular}
    \right. \nonumber
\end{align}
where $N$ is the number of identities in the total training set, and $\xi$ is a constant to encourage the model to be less confident on the training set. In this work, $\xi$ is set to 0.1.

We adopt both the identification loss and hetero-center triplet loss for each part-level feature, while only the hetero-center triplet loss $L_{hc\_tri}^{g}$ is for the final concatenated global features. Therefore, the final loss is,
\begin{align}
    L_{all}  =   L_{hc\_tri}^{g} + \sum_{i=1}^{p} \big(L_{id}^{i} +  \lambda L_{hc\_tri}^{i}\big), \label{eq:final_loss}
\end{align}
where $\lambda$ is a predefined tradeoff parameters.

\section{Experiments}
\label{sec:exp}
\begin{figure}
\centering
\includegraphics[width=9cm]{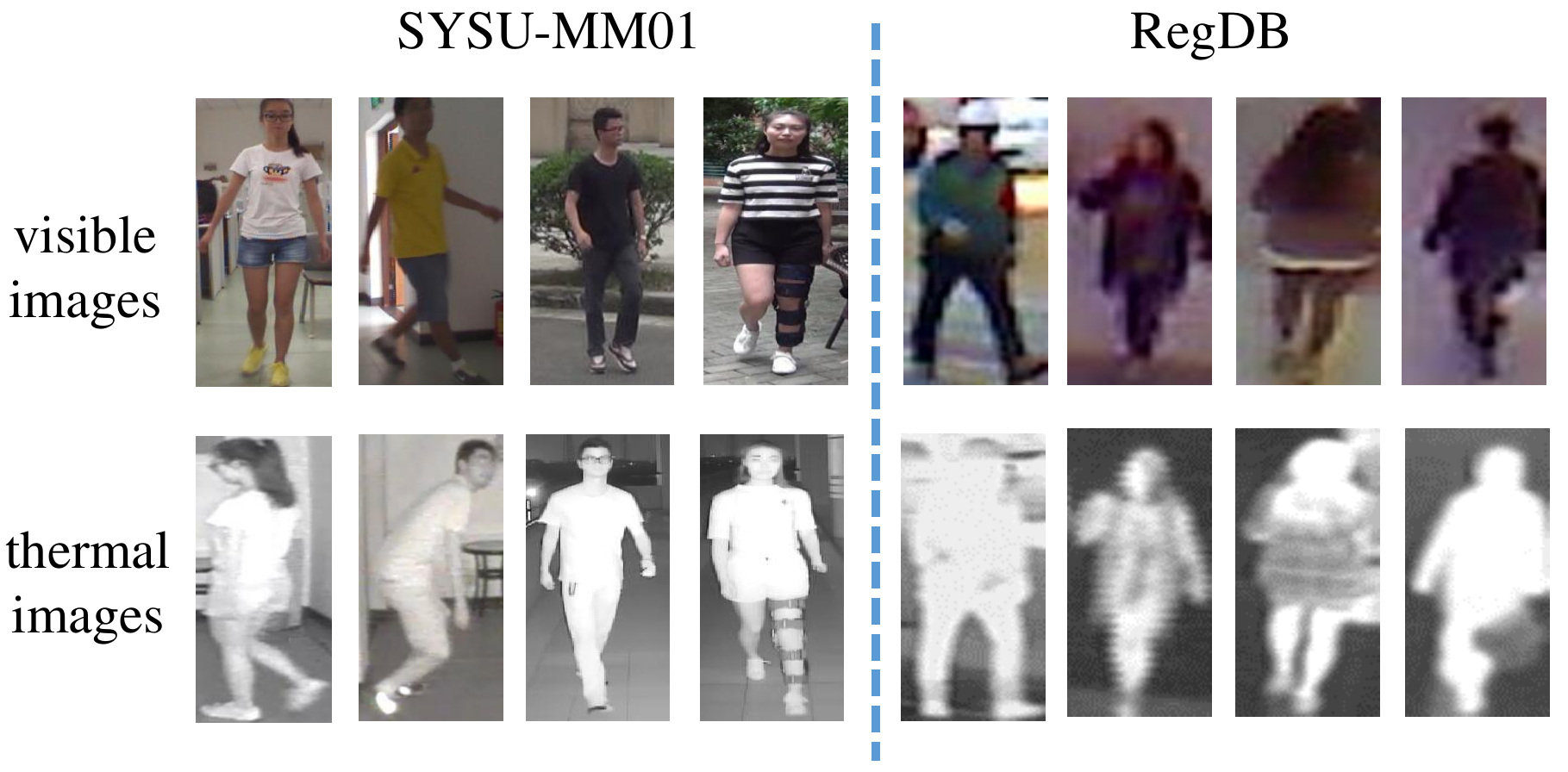}
\caption{Illustration of the visible-thermal images from two datasets, SYSU-MM01 \cite{wu2017rgb} and RegDB \cite{nguyen2017person}, for cross-modality person re-identification. The first row is the visible images, while the second is the thermal images. Each column contains images of the same person. }
\label{fig:image_illus}
\end{figure}

In this section, we evaluate the effectiveness of our proposed methods for extracting the person features for VT Re-ID tasks on two public datasets, RegDB \cite{nguyen2017person} and  SYSU-MM01 \cite{wu2017rgb}. The example images are shown in Fig. \ref{fig:image_illus}.
\subsection{Experimental settings}
\label{ssec:settings}

\subsubsection{Datasets and settings}
SYSU-MM01 \cite{wu2017rgb} is a large-scale dataset collected by 6 cameras, including 4 visible and 2 infrared cameras, captured in the SYSU campus. Some cameras are deployed in indoor environments and others are deployed in outdoor environments.  The training set contains 395 persons, including 22,258 visible images and 11,909 infrared images. The testing set contains another 96 persons, including 3,803 infrared images for the query and 301 randomly selected visible images as the gallery set.
In the \emph{all-search} mode, the gallery set contains all the visible images captured from all four visible cameras. In \emph{indoor-search} mode, the gallery set only contains the visible images captured by two indoor visible cameras. Generally, the all-search mode is more challenging than the indoor-search mode. We follow existing methods to perform 10 trials of gallery set selection in the single-shot setting \cite{ye2018visible,tifs19vtreid}, and then report the average retrieval performance. A detailed description of the evaluation protocol can be found in \cite{wu2017rgb}.

RegDB \cite{nguyen2017person} was constructed by dual-camera (one visible and one thermal camera) systems, and includes 412 persons. For each person, 10 visible images were captured by a visible camera, and 10 thermal images are obtained by a thermal camera. We follow the evaluation protocol in \cite{ye2018hierarchical} and \cite{ye2018visible}, where the dataset is randomly split into two halves, one for training and the other for testing. For testing, the images from one modality (default is thermal) were used as the gallery set while those from the other modality (default is visible) were used as the probe set. The procedure is repeated for 10 trials to achieve statistically stable results, recording the mean values.

\subsubsection{Evaluation Metrics}
Following existing works, cumulative matching characteristics (CMC), mean average precision (mAP) and the mean inverse negative penalty (mINP) are adopted as the evaluation metrics. CMC (rank-r accuracy) measures the probability of a correct cross-modality person image occurring in the top-r retrieved results. mAP measures the retrieval performance when multiple matching images occur in the gallery set. Moreover, mINP considers the hardest correct match that determines the workload of inspectors \cite{Ye2020DeepLF}.
Note that all the person features are first $L2$ normalized for testing.

\subsubsection{Implementation details}
The implementation\footnote{\url{https://github.com/hijune6/Hetero-center-triplet-loss-for-VT-Re-ID}} of our method is with the Pytorch framework. Following the existing person Re-ID works, the ResNet50 model is adopted as the backbone network for a fair comparison, and the pre-trained ImageNet parameters are adopted for the network initialization. Specifically, the stride of the last convolutional block is changed from 2 to 1 to obtain fine-grained feature maps with large body size. In the training phase, the input images are resized to $288 \times 144$ and padded with 10, then randomly left-right flipped and cropped to $288 \times 144$ for data augmentation.
We adopt the stochastic gradient descent (SGD) optimizer for optimization, and the momentum parameter is set to 0.9. We set the initial learning rate as 0.1 for both datasets. The warmup learning rate strategy is applied to bootstrap the network to enhance performance. The learning rate ($lr$) at epoch $t$ is computed as follows,
\begin{align}
    lr(t) =
    \left\{
    \begin{tabular}{cl}
        $ 0.1 \times \frac{t+1}{10}$, & $0 \leq t < 10$ \\
        $0.1$, & $10 \leq t < 20$ \\
        $0.01$, & $20 \leq t < 50$ \\
        $0.001$, & $50 \leq t $
    \end{tabular}.
    \right.
\end{align}

We set the predefined margin $\rho = 0.3$ for all triplet losses.
For the $PK$ sampling strategy, we set $P = 8$, $K=4$ for the RegDB dataset, and $P=6$, $K=8$ for the SYSU-MM01 dataset.
For the tradeoff parameter, we set $\lambda = 2.0$ for the RegDB dataset, and $\lambda = 1.0$ for the SYSU-MM01 dataset.
The dimension of part-level feature $d$ is set to $256$, and the number of part-level stripes $p$ is set to $6$.

\subsection{Ablation experiments}
\label{ssec:ablation}
We evaluate the effectiveness of our proposed method, including three components, two-stream backbone network, part-level feature learning and hetero-center triplet loss.\footnote{Note that to simply show the effectiveness of different components, during the ablation experiments, we only reported one trial experimental results, rather than the mean results of 10 trials.}

\subsubsection{Two-stream backbone network setting}
\label{sssec:banckbone}
\begin{figure}
\centering
\includegraphics[width=9cm]{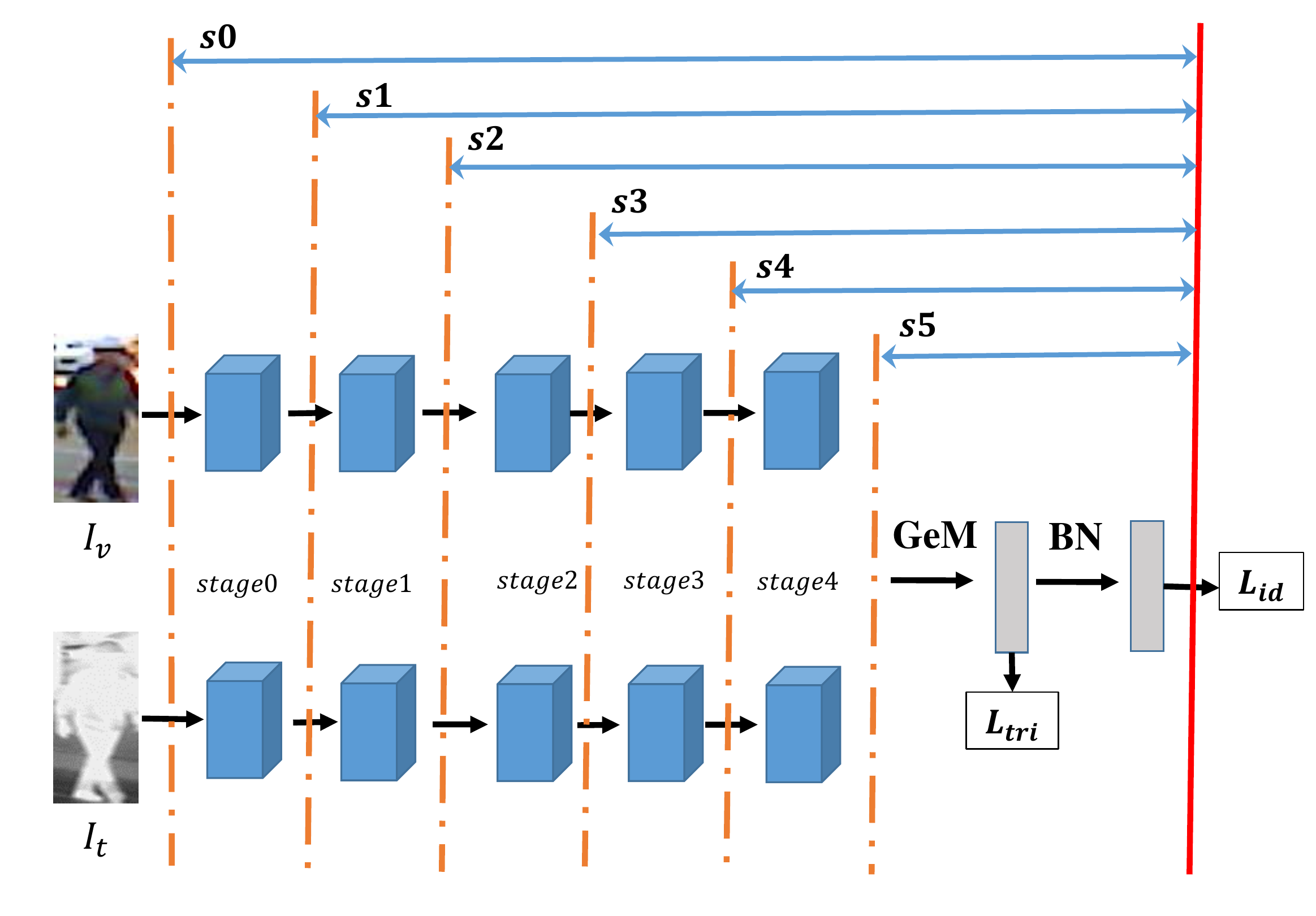}
\caption{Illustration of the baseline network, mainly illustrating how to split the ResNet50 model to set the two-stream backbone network. $si, i=\{0,1,2,3,4,5\}$ denotes that the modality-shared feature embedding block with parameter sharing, starting from $i^{th}$ $stage$. Then, the batch normalization neck with triplet loss and identification loss is adopted to train the network.}
\label{fig:net_share}
\end{figure}

As analyzed in Sec. \ref{ssec:backbone}, the key point of the two-stream backbone network setting is how to split the well-designed CNN model to construct a modality-specific feature extractor with independent parameters and modality-shared feature embedding with shared parameters.
Based on the AGW baseline \cite{Ye2020DeepLF} which is designed on top of BagTricks \cite{Luo2019ASB}, we optionally build the following baseline network with the ResNet50 model.
As shown in Fig. \ref{fig:net_share}, the 3D feature maps outputted from the two-stream backbone network are pooled by the generalized-mean pooling (GeM) layer to obtain the 2D feature vector. Then the batch normalization (BN) neck is adopted to train the network, where triplet loss (Eq. (\ref{eq:loss_bh})) is first utilized on the 2D feature vector, and then the identification loss is sequentially utilized on the batch normalized feature vector.

\begin{table}
\caption{The results of different splits of the backbone to form the two-stream network. $si, i=\{0,1,2,3,4,5\}$ denotes that the modality-shared feature embedding block with parameters sharing starts from $i^{th}$ $stage$. Re-identification rates at rank r, mAP and mINP (\%).}
\label{tab:net_share}
  \centering
  \begin{tabular}{l|c|c|c|c|c|c}
    \toprule[2pt]
    splits  & r = 1 & r = 5 & r = 10 & r = 20 & mAP & mINP \\ \toprule[2pt]
    \multicolumn{6}{c}{RegDB} \\ \hline
    $s0$ & \textbf{77.52} & 86.50 &  90.49 & 93.50 & 69.79 & 54.58 \\
    $s1$ & 76.94 & 85.68 & 89.71  & 93.88 & 69.36 & \textbf{54.82} \\
    $s2$ & 77.14 & \textbf{87.33} & \textbf{91.94}  & \textbf{95.19} & \textbf{69.82} & 54.62 \\
    $s3$ & 76.99 & 87.23 & 91.21  & 94.51 & 69.51 & 53.74 \\
    $s4$ & 64.95 & 78.30 & 85.00  & 90.49 & 60.98 & 48.62 \\
    $s5$ & 48.93 & 61.99 & 71.50  & 80.44 & 48.30 & 37.66 \\  \toprule[2pt]
    \multicolumn{6}{c}{SYSU-MM01}\\ \hline
    $s0$ & 54.38 & 80.78 & \textbf{88.96}  & \textbf{95.06} & 52.18 & 38.57 \\
    $s1$ & 54.48 & 80.38 & 88.61  & 94.61 & 52.67 & 39.19 \\
    $s2$ & \textbf{57.09} & \textbf{81.78} & 88.80  & 94.61 & \textbf{54.99} & \textbf{41.26} \\
    $s3$ & 52.20 & 78.23 & 86.83  & 93.06 & 51.43 & 39.49 \\
    $s4$ & 45.23 & 71.50 & 79.65  & 88.51 & 45.43 & 33.25 \\
    $s5$ & 37.81 & 69.18 & 79.88  & 87.96 & 39.40 & 27.68 \\ \toprule[2pt]
  \end{tabular}
\end{table}

The results of different backbone splits on RegDB and SYSU-MM01 datasets are listed in Table \ref{tab:net_share}, from which we can observe that.
\begin{enumerate}[a)]
  \item $s5$, without sharing any res-convolutional layers, obtains the worst performance on both the RegDB and SYSU-MM01 datasets, with large margins compared to other splits. $s5$ only shares the last fully connected layer to process the 1D feature vector without any person spatial structure information. It demonstrates the effectiveness of the 3D feature maps with person spatial structure information to describe a person.
  \item $s0$, sharing all the backbone networks without the modality-specific feature extractor, obtains good performances on both the RegDB and SYSU-MM01 datasets. $s0$ equally treats both visible and thermal person images, without focusing additionally on the color information of visible images, focusing on the spatial structure information of a person existing on both visible and thermal images. The results may demonstrate that the person spatial structure information is more important compared to the color information in VT cross-modality Re-ID.
  \item On the RegDB dataset, $s0$, $s1$, $s2$ and $s3$ achieve comparable performances. On the SYSU-MM01 dataset, $s2$ obtains much better Rank1, mAP and mINP results compared to $s0$ and $s1$. The different performances may be from the different settings of the two datasets. RegDB is collected by a dual-camera system, where the visible image and corresponding thermal image are well aligned. SYSU-MM01 is collected by 6 disjoint cameras deployed at different locations, where the visible image and corresponding infrared image have arbitrary poses and views. Therefore, SYSU-MM01 needs more modality-specific layers to extract the person spatial structure compared to RegDB.
  \item Overall, $s2$ can achieve the best performance, which only sets $stage0$ and $stage1$ as the modality-specific feature extractor with acceptable independent parameters.
\end{enumerate}

\begin{figure}
\centering
\includegraphics[width=8cm]{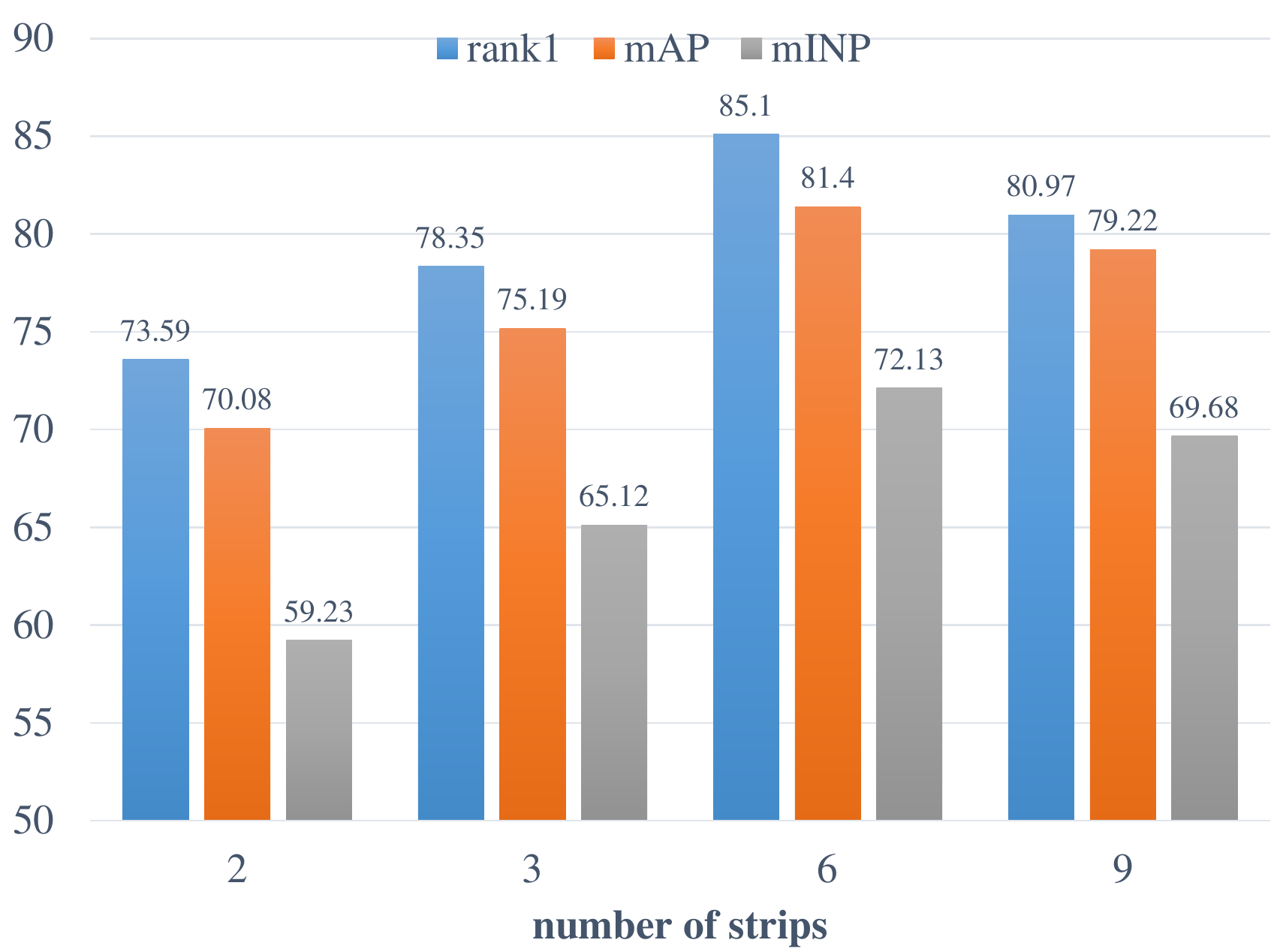} \\ (a) RegDB \\
\includegraphics[width=8cm]{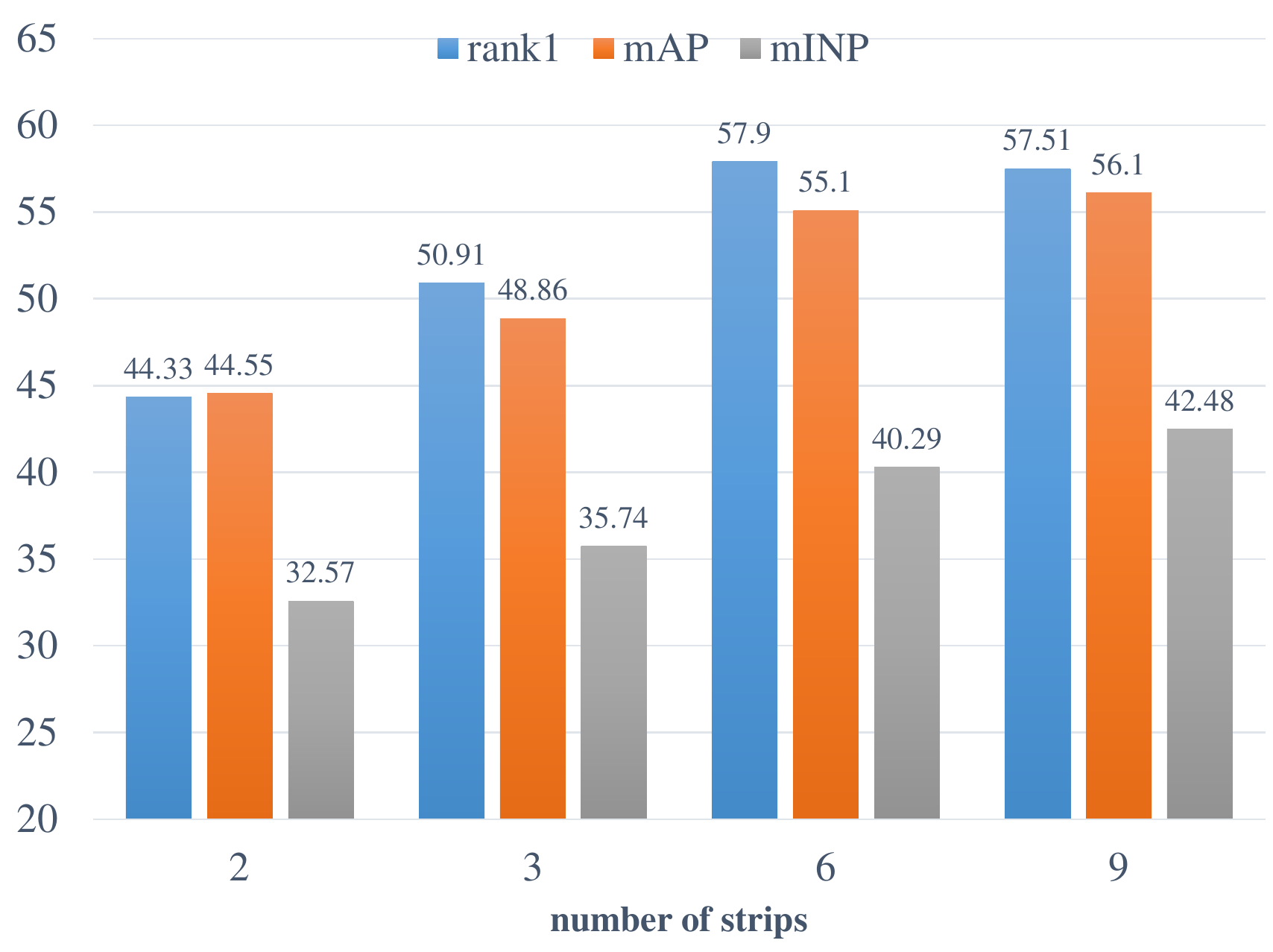} \\  (b) SYSU-MM01
\caption{The effect of partition strips $p$ on (a) RegDB and (b) SYSU-MM01 datasets. Re-identification rates of rank1, mAP and mINP (\%).}
\label{fig:strips}
\end{figure}

\subsubsection{Part-level feature learning}
\label{sssec:pfl}
To evaluate the effectiveness of part-level feature learning compared to global feature learning, we add the uniform partitioning strategy between the two-stream backbone network and the loss layer, as shown in Fig. \ref{fig:framework}.
Based on the above experimental results, we adopt the $s2$ split as the two-stream backbone network, still with the supervision of identification loss and triplet loss.

Three points are considered: 1) the number of partition strips $p$, 2) the generalized-mean pooling (GeM) layer instead of the traditional average pooling layer or max-pooling layer, and 3) the dimension of each part-level feature $d$ corresponding to the output channel number of $1 \times 1$ \textbf{Conv}.

\textbf{The effect of partition strips.} The number of partition strips determines the granularity of a person's local feature. Fig. \ref{fig:strips} shows the results of different partition strips $p$ on the RegDB and SYSU-MM01 datasets. We observe that $p=6$ is the best setting for partition strips to extract the local person feature.

\begin{table}
\caption{The results of different pooling methods on RegDB and SYSU-MM01 datasets, including generalized-mean pooling (GeM), average pooling (Mean) and max pooling (Max). Re-identification rates of rank1, mAP and mINP (\%).}
\label{tab:pool}
  \centering
  \begin{tabular}{l|c|c|c|c|c|c}
   \toprule[2pt]
    & \multicolumn{3}{c|}{RegDB} & \multicolumn{3}{c}{SYSU-MM01}\\ \hline
    Methods & rank1 & mAP & mINP & rank1 & mAP & mINP \\ \toprule[1pt]
    GeM & \textbf{85.10} & \textbf{81.40} & \textbf{72.13} & \textbf{57.90} & \textbf{55.10} & 40.29 \\
    Mean &  76.75 & 76.08 & 68.36 & 52.09 & 49.92 & 35.88 \\
    Max & 84.22 & 79.75 & 69.04 & 56.74 & 54.88 & \textbf{40.72} \\ \toprule[2pt]
  \end{tabular}
\end{table}

\textbf{The effect of GeM.} This subsection verifies the effectiveness of the generalized-mean pooling (GeM) method compared to the traditional average pooling (Mean) and max-pooling (Max) methods. Table \ref{tab:pool} lists the results of different pooling methods on the RegDB and SYSU-MM01 datasets. We observe that max-pooling performs better than average pooling, while the generalized-mean pooling method performs the best.

\textbf{The effect of part-level feature dimension.} This subsection shows the effect of part-level feature dimension $d$, corresponding to the output channel number of $1 \times 1$ Conv in Fig. \ref{fig:framework}. The final dimension of the person feature is the product of the part-level feature dimension $d$ and the number of partition strips $p$. Table \ref{tab:dimension} lists the results of different dimensions of each part-level feature $d$ on the RegDB and SYSU-MM01 datasets. We find that on SYSU-MM01 dataset, $d = 256$ performs the best, while on the RegDB dataset $d=512$ performs the best under the rank1 and mAP criteria, and $d=256$ achieves the best performance under the mINP criterion.
considering both the performance and final person feature dimension, we set $d=256$ for both the RegDB and SYSU-MM01 datasets.

\begin{table}
\caption{The results of different dimensions of each part-level feature $d$ on the RegDB and SYSU-MM01 datasets. Re-identification rates of rank1, mAP and mINP (\%).}
\label{tab:dimension}
  \centering
  \begin{tabular}{l|c|c|c|c|c|c}
   \toprule[2pt]
    & \multicolumn{3}{c|}{RegDB} & \multicolumn{3}{c}{SYSU-MM01}\\ \hline
    $d$ & rank1 & mAP & mINP & rank1 & mAP & mINP \\ \toprule[1pt]
    128 & 82.72 & 79.66 & 69.96 & 55.25 & 53.49 & 39.36\\
    256 & 85.10 & 81.40 & \textbf{72.13} & \textbf{57.90} & \textbf{55.10} & \textbf{40.29} \\
    512 & \textbf{86.99} & \textbf{82.02} & 71.66 & 57.17 & 53.89 & 38.80\\ \toprule[2pt]
  \end{tabular}
\end{table}

\subsubsection{Hetero-center based triplet loss}
\label{sssec:ctri}

In this subsection, we verify the effectiveness of our proposed hetero-center triplet loss $L_{hc\_tri}$ from two aspects. On the one hand, $L_{hc\_tri}$ is compared to traditional triplet loss $L_{bh\_tri}$ to demonstrate the effectiveness of \emph{anchor center to all the other centers} compared to \emph{anchor to all the other samples}. On the other hand, $L_{hc\_tri}$ is compared to the learned center loss $L_{lc}$ and hetero-center loss $L_{hc}$ to demonstrate the effectiveness of constraining both the inter-class separability and intra-class compactness.

\textbf{$L_{hc\_tri}$ vs. $L_{bh\_tri}$.} We conducted experiments under the framework shown in Fig. \ref{fig:framework} with different triplet losses, $L_{hc\_tri}$ and $L_{bh\_tri}$, fine-tuning the tradeoff parameter $\lambda$ in Eq. (\ref{eq:final_loss}). The results are listed in Table \ref{tab:tri}.
From the table, we can observe that 1) $L_{hc\_tri}$ outperforms $L_{bh\_tri}$ on both RegDB and SYSU-MM01 datasets, demonstrating the effectiveness of \emph{anchor center to all the other centers} compared to \emph{anchor to all the other samples}.
2) With the final loss Eq. (\ref{eq:final_loss}), those nonconvergent cases on the SYSU-MM01 dataset may show that the \emph{anchor to all the other samples} of $L_{bh\_tri}$ is truly a strict constraint, demonstrating the effectiveness of the \emph{anchor center to all the other centers} relaxation operation.

\begin{table}
\caption{The experimental results of $L_{hc\_tri}$ is compared to traditional triplet loss $L_{bh\_tri}$ on the RegDB and SYSU-MM01 datasets. Re-identification rates of rank1, mAP and mINP (\%). Note that $L_{bh\_tri}$ loss is not converged on SYSU-MM01 dataset when $\lambda \geq 0.5$.}
\label{tab:tri}
  \centering
  \begin{tabular}{l|c|c|c|c|c|c|c}
   \toprule[2pt]
    \multicolumn{2}{c|}{} & \multicolumn{3}{c|}{RegDB} & \multicolumn{3}{c}{SYSU-MM01}\\ \hline
    Loss & $\lambda$ & rank1 & mAP & mINP & rank1 & mAP & mINP \\ \toprule[1pt]
    \multirow{4}{*}{$L_{bh\_tri}$}& 0.1 & 80.68 & 75.35 & 64.50 & 55.30 & 54.21 & 41.09\\
    & 0.5 & 82.43 & 78.93 & 69.22 & - & - & - \\
    & 1.0 & 85.10 & 81.40 & 72.13 & - & - & - \\
    & 1.5 & 72.33 & 69.67 & 60.55 & - & - & -  \\ \hline \hline
    \multirow{6}{*}{$L_{hc\_tri}$}& 0.1 &  80.73 & 75.08 & 64.14 & 57.53 & 54.68 & 40.22 \\
    &0.5 & 87.96 & 81.97 & 71.87 & 60.43 & 56.41 & \textbf{40.56} \\
    &1.0 & 85.24 & 81.76 & \textbf{72.33} &  \textbf{61.95} &\textbf{ 57.25} & 40.44 \\
    &1.5 & 90.63 & 83.64 & 71.21 & 57.45 & 53.01 & 36.93 \\
    &2.0 & \textbf{92.48} & \textbf{84.41} & 71.53 & - & - & - \\
    &3.0 & 88.93 & 78.64 & 62.22 & - & - & - \\ \toprule[2pt]
  \end{tabular}
\end{table}

\begin{table}
\scriptsize
\caption{The results of different center-based losses on the RegDB and SYSU-MM01 datasets, including the learned center loss $L_{lc}$, hetero-center loss $L_{hc}$, and our proposed hetero-center triplet loss $L_{hc\_tri}$. Re-identification rates of rank1, mAP and mINP (\%).}
\label{tab:centerloss}
  \centering
  \begin{tabular}{l|c|c|c|c|c|c|c}
   \toprule[2pt]
    \multicolumn{2}{c|}{} & \multicolumn{3}{c|}{RegDB} & \multicolumn{3}{c}{SYSU-MM01}\\ \hline
    Network & Loss & rank1 & mAP & mINP & rank1 & mAP & mINP \\ \toprule[1pt]
    \multirow{3}{*}{Global-level}& $L_{lc}$ &  44.76 & 40.60 & 27.52 & 52.67 & 50.48 & 36.84\\
    & $L_{hc}$ & 52.96 & 44.52 & 27.05 & 51.30 & 48.12 & 33.73 \\
    & $L_{hc\_tri}$ & \textbf{79.22} & \textbf{68.35} & \textbf{48.87} &  \textbf{57.61} & \textbf{53.03} & \textbf{37.22}  \\ \hline \hline
    & $L_{lc}$ &  67.38 & 64.30 & 54.83 &  46.02 & 47.55 & 36.70  \\
    Part-level & $L_{hc}$ & 85.34 & 80.83 & 70.46 &  47.83 & 46.22 & 32.48 \\
    (ours)&$L_{hc\_tri}$ & \textbf{92.48} & \textbf{84.41} & \textbf{71.53} & \textbf{61.95} & \textbf{57.25} & \textbf{40.44} \\ \toprule[2pt]
  \end{tabular}
\end{table}

\textbf{$L_{hc\_tri}$ vs. $L_{lc}$ and $L_{hc}$.} We conducted experiments with different center-based losses, including the learned center loss $L_{lc}$, hetero-center loss $L_{hc}$, and our proposed hetero-center triplet loss $L_{hc\_tri}$. The network is in two frameworks: the baseline network that extracts the global person features (Sec. \ref{sssec:banckbone}) and the part-level local feature learning network (Sec. \ref{sssec:pfl}). The results are listed in Table \ref{tab:centerloss}. We can observe that in both networks, $L_{hc\_tri}$ outperforms $L_{lc}$ and $L_{hc}$ with large margins. It demonstrates the effectiveness of our proposed $L_{hc\_tri}$ concentrating on both the inter-class separability and intra-class compactness, compared to $L_{lc}$ and $L_{hc}$ which only focus on intra-class cross-modality compactness, ignoring the inter-class separability for both the intra- and inter-modality. It is also illustrated in Fig. \ref{fig:visualization} through visualizing the features extracted by the baseline model with different center-based losses.

\subsubsection{Ablation summarization}
\label{sssec:ablasum}
Moreover, to show the effect of every component, we also summarize the corresponding ablation study in Table \ref{tab:ablation}, whose results are copied from Tables \ref{tab:net_share}, \ref{tab:tri} and \ref{tab:centerloss}. It shows that when the component works alone, the performance on the two datasets is different and is not always improved. However, the combination of three components can greatly boost the performance on both datasets.

\begin{table}
\scriptsize
\caption{The ablation study of different components: weight sharing ($ws$), hetero-center triplet loss ($L_{hc\_tri}$) and part-level feature learning ($plf$). Re-identification rates of rank1, mAP and mINP (\%).}
\label{tab:ablation}
  \centering
  \begin{tabular}{c|c|c|c|c|c|c|c|c}
   \toprule[2pt]
    \multicolumn{3}{c|}{} & \multicolumn{3}{c|}{RegDB} & \multicolumn{3}{c}{SYSU-MM01}\\ \hline
    $ws$ & $L_{hc\_tri}$ & $plf$ & rank1 & mAP & mINP & rank1 & mAP & mINP \\ \toprule[1pt]
    $\times$ & $\times$ & $\times$ & 77.52 & 69.79 & 54.58 & 54.38 & 52.18 & 38.57 \\
    $\checkmark$ & $\times$ & $\times$ & 77.14 & 69.82 & 54.62 & 57.09 & 54.99 & 41.26 \\
    $\checkmark$ & $\checkmark$ & $\times$ & 79.22 & 68.35 & 48.87 & 57.61 & 53.03 & 37.22 \\
    $\checkmark$ & $\times$ & $\checkmark$ & 85.10 & 81.40 & \textbf{72.13} & 55.30 & 54.21 & \textbf{41.09} \\
    $\checkmark$ & $\checkmark$ & $\checkmark$ & \textbf{92.48} & \textbf{84.41} & 71.53 & \textbf{61.95} & \textbf{57.25} & 40.44 \\
     \toprule[2pt]
  \end{tabular}
\end{table}

\begin{table}
\footnotesize
\caption{Comparison to the state-of-the-art methods on the RegDB datasets in visible $\rightarrow$ thermal and thermal $\rightarrow$ visible query settings. Re-identification rates at rank r, mAP and mINP (\%).}
\label{tab:sota_regdb}
  \centering
  \begin{tabular}{l|c|c|c|c|c|c}
    \toprule[2pt]
      Methods & Venue &   r = 1  & r = 10 & r = 20 & mAP & mINP    \\ \toprule[1pt]
      \multicolumn{7}{c}{Visible $\rightarrow$ Thermal} \\ \toprule[1pt]
      Zero-Pad \cite{wu2017rgb}& ICCV17 &  17.75 & 34.21 & 44.35 & 18.90 & -  \\
      HCML \cite{ye2018hierarchical} & AAAI18 & 24.44 & 47.53 & 56.78& 20.80 & -  \\ \hline
      HSME \cite{hao2019hsme} & AAAI19  & 50.85 & 73.36 & 81.66 & 47.00 & -  \\
      D$^2$RL \cite{wang2019learning1}& CVPR19  & 43.40 &66.10 &76.30 &44.10 & -  \\
      MAC \cite{Ye2019ModalityawareCL} & MM19 & 36.43 & 62.36 & 71.63 & 37.03 & -  \\
      AliGAN \cite{wang2019rgb} & ICCV19 & 57.90 & - & - & 53.60 & -  \\
      DFE \cite{Hao2019DualalignmentFE} &  MM19 & 70.13 & 86.32 & 91.96 & 69.14 & -  \\ \hline
      eBDTR \cite{Ye2020DeepLF} & TIFS20 & 34.62 & 58.96 & 68.72 & 33.46 &-  \\
      MSR \cite{Feng2020LearningMR} & TIP20 & 48.43 & 70.32 & 79.95 & 48.67 &-  \\
      JSIA \cite{Wang2020CrossModalityPG} & AAAI20 & 48.50 & - & - & 48.90 &-  \\
      EDFL \cite{liu2020enhancing} & Neuro20 & 52.58 & 72.10 & 81.47 & 52.98 & -  \\
      XIV \cite{Li2020InfraredVisibleCP} & AAAI20 & 62.21 & 83.13 & 91.72 & 60.18 &-  \\
      CDP \cite{Fan2020CrossSpectrumDP} & Arxiv20 & 65.00 & 83.50 & 89.60 & 62.70 &-  \\
      expAT \cite{Ye2020BidirectionalEA} & Arxiv20 & 66.48 & - & - & 67.31 &-  \\
      CMSP \cite{wu2020rgb} & IJCV20 & 65.07 & 83.71 & - & 64.50 &-  \\
      Hi-CMD \cite{choi2020hi} & CVPR20 & 70.93 & 86.39 & - & 66.04 & -   \\
      HAT \cite{ye2018vipr} & TIFS20 & 71.83 & 87.16 & 92.16 & 67.56 & - \\
      cmSSFT \cite{lu2020cross} & CVPR20 & 72.30 & - & - & 72.90 &-  \\
      MPMN \cite{Wang2020MPMN} & TMM20 & 86.56 & 96.68 & 98.28 & 82.91 & - \\
      AGW \cite{Ye2020DeepLF} & Arxiv20 & 70.05 & - & - & 66.37 & 50.19  \\ \hline
      ours & - & \textbf{91.05}  & \textbf{97.16} & \textbf{98.57} & \textbf{83.28} & \textbf{68.84} \\ \toprule[1pt] \toprule[1pt]
      \multicolumn{7}{c}{ Thermal $\rightarrow$ Visible} \\ \toprule[1pt]
      Zero-Pad \cite{wu2017rgb}& ICCV17 & 16.63 & 34.68 & 44.25 & 17.82 & - \\
      HCML\cite{ye2018hierarchical} & AAAI18 & 21.70 & 45.02 & 55.58 & 22.24 & -\\
      eBDTR \cite{Ye2020DeepLF} & TIFS20 & 34.21 & 58.74 & 68.64 & 32.49 & - \\
      MAC \cite{Ye2019ModalityawareCL} & MM19 & 36.20 & 61.68 & 70.99 & 39.23 & - \\
      HSME \cite{hao2019hsme} & AAAI19  & 50.15 & 72.40 & 81.07 & 46.16 & - \\
      EDFL \cite{liu2020enhancing} & Neuro20 & 51.89 & 72.09 & 81.04 & 52.13 & - \\
      AliGAN \cite{wang2019rgb} & ICCV19 & 56.30 & - & - & 53.40 & - \\
      expAT \cite{Ye2020BidirectionalEA} & Arxiv20 & 67.45 & - & - & 66.51 & - \\
      MPMN \cite{Wang2020MPMN} & TMM20 & 84.62 & 95.51 & 97.33 & 79.49 & - \\ \hline
      ours & - & \textbf{89.30}  & \textbf{96.41} & \textbf{98.16} & \textbf{81.46} & \textbf{64.81} \\
      \toprule[2pt]
  \end{tabular}
\end{table}

\begin{table*}
\caption{Comparison to the state-of-the-art methods on the SYSU-MM01 datasets. Re-identification rates at rank r, mAP and mINP (\%).}
\label{tab:sota_sysu}
  \centering
  \begin{tabular}{l|c|c|c|c|c|c||c|c|c|c|c}
    \toprule[2pt]
    \multicolumn{2}{c|}{}  & \multicolumn{5}{c||}{\emph{All search}} & \multicolumn{5}{c}{\emph{Indoor search}} \\ \hline
      Methods & Venue &   r = 1  & r = 10 & r = 20 & mAP & mINP  &  r = 1 &  r = 10 & r = 20 & mAP & mINP     \\ \toprule[1pt]
      Zero-Pad \cite{wu2017rgb}& ICCV17 & 14.80 & 54.12 & 71.33 & 15.95 & - & 20.58 & 68.38 & 85.79 & 26.92 &-  \\
      cmGAN \cite{dai2018cross} & IJCAI18 & 26.97 & 67.51 & 80.56 & 27.80 & - & 31.63 & 77.23 & 89.18 & 42.19 &-  \\
      HCML \cite{ye2018hierarchical} & AAAI18 & 14.32 & 53.16 & 69.17 & 16.16 &- & 24.52 & 73.25 & 86.73 & 30.08 &- \\ \hline
      HSME \cite{hao2019hsme} & AAAI19  & 20.68 & 62.74 & 77.95 & 23.12 &- & - & - & - & - & - \\
      D$^2$RL \cite{wang2019learning1}& CVPR19  & 28.90 & 70.60 & 82.40 & 29.20 &- & - & - & - & - & - \\
      MAC \cite{Ye2019ModalityawareCL} & MM19 & 33.26 & 79.04 & 90.09 & 36.22 &- & 36.43 & 62.36 & 71.63 & 37.03 &- \\
      AliGAN \cite{wang2019rgb} & ICCV19 & 42.40 & 85.00 & 93.70 & 40.70 &- & 45.90 & 87.60 & 94.40 & 54.30 &- \\
      HPILN \cite{zhao2019hpiln} & TIP19 & 41.36 & 84.78 & 94.51 & 42.95 &- & 45.77 & 91.82 & 98.46 & 56.52 &- \\
      DFE \cite{Hao2019DualalignmentFE} &  MM19 & 48.71 & 88.86 & 95.27 & 48.59 &- & 52.25 & 89.86 & 95.85 & 59.68 &-  \\ \hline
      Hi-CMD \cite{choi2020hi} & CVPR20 & 34.94 & 77.58 & - & 35.94 & -  &  - & - & - & - & - \\
      EDFL \cite{liu2020enhancing} & Neuro20 & 36.94 & 85.42 & 93.22 & 40.77 & - &  - & - & - & - & - \\
      CDP \cite{Fan2020CrossSpectrumDP} & Arxiv20 & 38.00 & 82.30 & 91.70 & 38.40 &- &  - & - & - & - & - \\
      expAT \cite{Ye2020BidirectionalEA} & Arxiv20 & 38.57 & 76.64 & 86.39 & 38.61 &- &  - & - & - & - & - \\
      XIV \cite{Li2020InfraredVisibleCP} & AAAI20 & 49.92 & 89.79 & 95.96 & 50.73 &- &  - & - & - & - & - \\
      eBDTR \cite{Ye2020DeepLF} & TIFS20 & 27.82 & 67.34 & 81.34 & 28.42 &- & 32.46 & 77.42 & 89.62 & 42.46 &- \\
      MSR \cite{Feng2020LearningMR} & TIP20 & 37.35 & 83.40 & 93.34 & 38.11 &- & 39.64 & 89.29 & 97.66 & 50.88 &- \\
      JSIA \cite{Wang2020CrossModalityPG} & AAAI20 & 38.10 & 80.70 & 89.90 & 36.90 &- & 43.80 & 86.20 & 94.20 & 52.90 &- \\
      CMSP \cite{wu2020rgb} & IJCV20 & 43.56  & 86.25 & - & 44.98 &- & 48.62  & 89.50 & -  & 57.50 &- \\
      Attri \cite{zhang20deepfl} & JEI20 & 47.14 & 87.93 & 94.45 & 47.08 & -& 48.03 & 88.13 & 95.14 & 56.84 &- \\
      HAT \cite{ye2018vipr} & TIFS20 & 55.29 & 92.14 & \textbf{97.36} & 53.89 & -& 62.10 & \textbf{95.75} & \textbf{99.20} & \textbf{69.37} &- \\
      HC \cite{zhu2019hetero} & Neuro20 & 56.96 & 91.50 & 96.82 & 54.95 &- & 59.74 & 92.07 & 96.22 & 64.91 &- \\
      AGW \cite{Ye2020DeepLF} & Arxiv20 & 47.50 & - & - & 47.65 & 35.30 & 54.17 & - & -& 62.97 & 59.23 \\ \hline
      ours & - & \textbf{61.68}  & \textbf{93.10} & 97.17 & \textbf{57.51} & \textbf{39.54} & \textbf{63.41} & 91.69 & 95.28 & 68.17 & \textbf{64.26} \\
      \toprule[2pt]
  \end{tabular}
\end{table*}

\subsection{Comparison to the state-of-the-art}
\label{ssec:sota}

This section compares the state-of-the-art VT Re-ID methods. The results on the RegDB and SYSU-MM01 datasets are listed in Tables \ref{tab:sota_regdb} and \ref{tab:sota_sysu}, respectively. \footnote{Note that in this subsection, we reported the mean results of 10 trials following the standard dataset settings.}

The experiments on the RegDB dataset (Table \ref{tab:sota_regdb}) demonstrate that our proposed method obtains the best performance in both query settings, always by large margins. We set a new baseline for this dataset, achieving superior performance rank1/mAP/mINP 91.05\%/83.28\%/68.84\% for visible $\rightarrow$ thermal query setting. The experiments suggest that our proposed method can learn better cross-modality sharing features by well designing the two-stream parameter-sharing network, learning the part-level local person features, and computing the triplet loss on heterogeneous centers from different modalities.

The experiments on the SYSU-MM01 dataset (Table \ref{tab:sota_sysu}) show that our proposed method can achieve comparable performance compared to the current state-of-the-art results obtained by HAT \cite{ye2018vipr}, and outperforms all the other comparison methods. However, in the more challenging mode \emph{all-search}, our method performs much better than HAT \cite{ye2018vipr} in the two key criteria rank1/mAP, 61.68\%/57.51\% vs. 55.29\%/53.89\%.

Compared to AliGAN \cite{wang2019rgb}, D$^2$RL \cite{wang2019learning1}and Hi-CMD \cite{choi2020hi}, our method achieves much better performance on both datasets, and does not need the sophisticated cross-modality image translation operation. Our method also does not require complicated adversarial learning with many tricks, which is always difficult for training.

\section{Conclusions}
This paper aims to enhance the discriminative person feature learning through simple means for VT Re-ID.
On the one hand, we explore the parameter-sharing settings in the two-stream network. The experimental results show that the modality-sharing feature embedding network with some convolution blocks is an effective strategy, that could process the 3D shape feature maps with the spatial structure of a person.
On the other hand, we also propose the hetero-center triplet loss to improve the traditional triplet loss for VT Re-ID by replacing the comparison of the \emph{anchor to all the other samples} with the \emph{anchor center to all the other centers}. With part-level person feature learning, hetero-center triplet loss performs much better than traditional triplet loss.
The experimental results with remarkable improvements on two VT Re-ID datasets demonstrate the effectiveness of our proposed method compared to the current state-of-the-art methods. Our method with a simple but effective strategy can be a strong VT Re-ID baseline to boost future research with high quality.


\bibliographystyle{IEEEtrans}
\bibliography{reid}

\end{document}